\documentclass[letterpaper, 10 pt, conference]{ieeeconf}  % Comment this line out if you need a4paper

\usepackage{graphicx}
\usepackage{float}
\usepackage{amsmath}
\usepackage{threeparttable, tablefootnote}

\IEEEoverridecommandlockouts                              % This command is only needed if 
\floatstyle{ruled}
\newfloat{algorithm}{tbp}{loa}
\providecommand{\algorithmname}{Algorithm}
\floatname{algorithm}{\protect\algorithmname}

\usepackage{subfigure}

\let\vec\boldvec

\DeclareMathOperator*{\argmax}{\textbf{argmax}}
\DeclareMathOperator*{\argmin}{\textbf{argmin}}
\usepackage{algpseudocode}
\usepackage{epstopdf}
\usepackage{multicol}
\usepackage{booktabs}

\usepackage[hidelinks]{hyperref}
\usepackage{amssymb}
\usepackage{bm}
\newcommand{\trsp}{{\!\scriptscriptstyle\top}}
\usepackage{arydshln}

\title{\LARGE \bf
A Linearly Constrained Nonparametric Framework for\\ Imitation Learning}
\author{Yanlong Huang and Darwin G. Caldwell % <-this % stops a space
\thanks{Yanlong Huang and Darwin G. Caldwell are with Department of Advanced Robotics, Istituto Italiano di Tecnologia,
	Via Morego 30, 16163 Genoa, Italy (\tt\small yanlong.huang@iit.it; darwin.caldwell@iit.it).}
}

\begin{document}

\maketitle
\thispagestyle{empty}
\pagestyle{empty}

%%%%%%%%%%%%%%%%%%%%%%%%%%%%%%%%%%%%%%%%%%%%%%%%%%%%%%%%%%%%%%%%%%%%%%%%%%%%%%%%
\begin{abstract}
In recent years, a myriad of advanced results have been reported in the community of imitation learning, ranging from parametric to non-parametric, probabilistic to non-probabilistic and Bayesian to frequentist approaches. Meanwhile, ample applications (e.g., grasping tasks and human-robot collaborations) 
further show the applicability of imitation learning in a wide range of domains. While numerous literature is dedicated to the learning of human skills in unconstrained environment, the problem of learning constrained motor skills, however, has not received equal attention yet. In fact, constrained skills exist widely in robotic systems. For instance, when a robot is demanded to write letters on a board, its end-effector trajectory must comply with the plane constraint from the board. 
In this paper, we aim to tackle the problem of imitation learning with linear constraints. Specifically, we propose to exploit the probabilistic properties of multiple demonstrations, and subsequently incorporate them into a linearly constrained optimization problem, which finally leads to a non-parametric solution. In addition, a connection between our framework and the classical model predictive control is provided.
Several examples including simulated writing and locomotion tasks are presented to show the effectiveness of our framework.

\end{abstract}

%%%%%%%%%%%%%%%%%%%%%%%%%%%%%%%%%%%%%%%%%%%%%%%%%%%%%%%%%%%%%%%%%%%%%%%%%%%%%%%%
\section{Introduction}
In the community of robot learning, a vital goal is to endow robots with the learning capabilities.
In this line, \emph{imitation learning} \cite{atkeson1997robot, schaal1999imitation}, also referred to as \emph{programming by demonstration} or \emph{learning from demonstration} \cite{billard2008robot},  emerges as an important research direction due to its nature and user-friendly features. Remarkably, many imitation learning approaches, such as dynamical movement primitives (DMP) \cite{ijspeert}, 
task-parameterized Gaussian mixture model (GMM) \cite{calinon2014task}  and kernelized movement primitives (KMP) \cite{huang2017}, have been developed. In addition, imitation learning has achieved great success in ample scenarios, e.g., reaching \cite{ude2010task}, grasping \cite{stulp2011learning} and striking \cite{huang2016} tasks.

However, most of the aforementioned works focus on learning human skills while the possible external or internal constraints are ignored. As we know, robot often encounter various constraints in dynamical environments. To take the wiping task as an example, the robot end-effector trajectory should always lie above the table since trajectories under the table are physically infeasible. Besides, the end-effector trajectory must obey the plane constraint of the table in order to clean the table successfully. It is also worth pointing out that, during the wiping process, robot joint trajectories must comply with the joint limits.

In order to cope with the constraints that are imposed to robots, a few approaches have been proposed. For example, the hard (i.e., equality) constraints were studied in \cite{howard2009novel, armesto2017learning} and joint limit avoidance\footnote{From an optimization perspective, joint limits can be viewed as a special case of inequality constraints.} was treated in \cite{ratliff,chaumette2001redundancy}. 
In contrast to these approaches, we aim at developing a generic framework which can be employed to learn and generalize human demonstrations to new situations, and meanwhile address a variety of constraints (i.e., equality and inequality constraints) that commonly arise in practice. In summary, the new framework should be capable of
\begin{enumerate}
	\item[(i)] inheriting the key features of imitation learning (e.g., learning multiple demonstrations, reproduction as well as adaptation towards via-/end- points in terms of position and velocity); 
	\item[(ii)] taking into account arbitrary linear equality (e.g., plane constraint) and inequality constraints (e.g., a linear combination of action components should be larger or smaller than a predefined value).
\end{enumerate}

To do so, we propose to built the linearly constrained imitation learning framework on the top of our previous work KMP \cite{huang2017}, since KMP provides us with most of the crucial features in imitation learning, including the learning and adaptation of human skills associated with time input and high-dimensional inputs. Specifically, we first exploit the probabilistic properties of multiple demonstrations in Section~\ref{sec:gmm}. Subsequently, we present the constrained framework in Section~\ref{sec:lc-kmp}. Since the classical model predictive control (MPC) can address the constrained trajectory optimization problem, we provide a connection between our framework and MPC in Section~\ref{sec:mpc}. Finally, we evaluate the proposed approach through several simulated examples in Section~\ref{sec:eva}.

\section{Exploiting Probabilistic Features of Multiple Demonstrations \label{sec:gmm}}
Following the spirit of probabilistic approaches \cite{huang2017,calinon2007,muhlig2009task,huang2018towards}, we model demonstrations from a probabilistic perspective. 
Let us assume that we have access to $M$ demonstrations $\vec{D}=\{\{t_{n,m}, \vec{\xi}_{n,m}, \dot{\vec{\xi}}_{n,m} \}_{n=1}^{N}\}_{m=1}^{M}$, where $N$ denotes the trajectory length, $\vec{\xi}\in \mathbb{R}^{\mathcal{O}}$ and $\dot{\vec{\xi}}$ respectively correspond to the output and its first-order derivative. For the sake of brevity, we write $\vec{\eta}=[\vec{\xi}^{\trsp}\,\dot{\vec{\xi}}^{\trsp}]^{\trsp}$. Then, we can employ GMM to model the joint probability distribution $\mathcal{P}(t,\vec{\eta})$, leading to
\begin{equation}
\mathcal{P}(t,\vec{\eta}) \sim \sum_{c=1}^{C} \pi_c \mathcal{N}(\vec{\mu}_c,\vec{\Sigma}_c)
\label{equ:gmm}
\end{equation}
with
$\pi_c$,
$\vec{\mu}_c=\left[\begin{matrix}
\vec{\mu}_{t,c} \\ \vec{\mu}_{\eta,c}
\end{matrix}\right]$ and
$\vec{\Sigma}_c=\left[\begin{matrix}
\vec{\Sigma}_{tt,c} & \vec{\Sigma}_{t\eta,c} \\ \vec{\Sigma}_{\eta t,c} & \vec{\Sigma}_{\eta \eta,c}
\end{matrix}\right]$ being the prior probability, mean and covariance of the $c$-th Gaussian component, respectively\footnote{Note that vector notations $\vec{u}_{t,c}$ and $\vec{\Sigma}_{tt,c}$ are used to represent scalars.}. Here, $C$ represents the number of Gaussian components. After that, we resort to Gaussian mixture regression\footnote{More details about GMM/GMR can be found in \cite{huang2017,cohn,calinon2016}.}  (GMR) to retrieve a \emph{probabilistic reference trajectory} $\{t_n,\hat{\vec{\eta}}_n\}_{n=1}^{N}$ with $\hat{\vec{\eta}}_n|t_n \sim \mathcal{N}(\hat{\vec{\mu}}_n,\hat{\vec{\Sigma}}_n)$, where the input sequence $\{t_n\}_{n=1}^{N}$ (e.g., equal interval sequence) spans the whole input space. 

We now denote
$\vec{D}_r=\{t_n,\hat{\vec{\mu}}_n,\hat{\vec{\Sigma}}_n\}_{n=1}^{N}$. In fact,  $\vec{D}_r$ encapsulates the probabilistic features of demonstrations $\vec{D}$, since it estimates the means and covariances of demonstrations over various key input points. In the next section, we will exploit $\vec{D}_r$ and propose the linearly constrained imitation learning framework in details.

\section{Linearly Constrained Nonparametric Framework for Imitation Learning \label{sec:lc-kmp}}

Let us first write $\vec{\eta}(t)$ using a parametric form\footnote{Similar parametric form was exploited in \cite{ijspeert,paraschos}.}, i.e., 

\begin{equation}
\vec{\eta}(t)\!\!=\!\!\left[\!\begin{array}{c}
\vec{\xi}(t) \\ \dot{\vec{\xi}}(t)
\end{array}\!\right]
=
\vec{\Theta}^{\trsp}(t)\vec{w}
\label{equ:para:traj}
\end{equation}
with
\begin{equation}
\vec{\Theta}(t)\!\!=\!\!\left[\begin{matrix} 
\vec{\varphi}(t) \!& \vec{0} \!& \cdots \!&\vec{0} \!& \dot{\vec{\varphi}}(t) \!& \vec{0} \!& \cdots \!&\vec{0} \\
\vec{0} \!& \vec{\varphi}(t) \!&  \cdots \!&\vec{0} \!&\vec{0} \!& \dot{\vec{\varphi}}(t) \!&  \cdots &\vec{0}\\
\vdots \!& \vdots \!&  \ddots \!& \vdots \!&\vdots \!& \vdots \!&  \ddots \!& \vdots\\
\vec{0} \!& \vec{0} \!&  \cdots \!& \vec{\varphi}(t) \!&\vec{0} \!& \vec{0} \!&  \cdots \!& \dot{\vec{\varphi}}(t)\\
\end{matrix}\right] \!,
\label{equ:basis:function:time}
\end{equation}
where $\vec{\varphi}(\cdot) \in \mathbb{R}^{B}$ represent a basis function vector and $\vec{w}\in \mathbb{R}^{B\mathcal{O}}$ denotes the \emph{unknown} parameter vector.

Formally, we formulate the problem of imitation learning with linear constraints as a constrained optimization problem
\begin{equation}
\begin{aligned}
\argmax_{\vec{w}}& \, \ \, \sum_{n=1}^{N}\mathcal{P}\left(\vec{\eta}(t_n)|\hat{\vec{\mu}}_n,\hat{\vec{\Sigma}}_n\right)\\
\quad\quad \textbf{s.t.} \, \ \,
&\left.
\begin{array}{cc}
\vec{g}_{n,1}^{\trsp}\vec{\eta}(t_n)& \geq c_{n,1} \\
\vec{g}_{n,2}^{\trsp}\vec{\eta}(t_n)&\geq c_{n,2} \\
\vdots&  \\
\vec{g}_{n,F}^{\trsp}\vec{\eta}(t_n)& \geq c_{n,F} \\
\end{array},
\right. \forall n\in\{1,2,\ldots,N\},
\end{aligned}
\label{equ:ini:problem}
\end{equation}
where we use $\vec{g}_{n,f} \in \mathbb{R}^{2\mathcal{O}}$ and $\vec{c}_{n,f} \in \mathbb{R}$ to characterize the $f$-th linear constraint over $\vec{\xi}(t_n)$ and $\dot{\vec{\xi}}(t_n)$, and $F$ to denote the number of constraints. 

With probability calculations and logarithm transformation, the constrained maximization problem in (\ref{equ:ini:problem}) can be rewritten as
\begin{equation}
\begin{aligned}
&\argmin_{\vec{w}} \!\! \sum_{n=1}^{N}\!\!\frac{1}{2}\!(\vec{\Theta}^{\trsp}(t_n)\vec{w}\!-\!\hat{\vec{\mu}}_n)^{\trsp}\! \hat{\vec{\Sigma}}_n^{-1}\!
\!(\vec{\Theta}^{\trsp}(t_n)\vec{w}\!-\!\hat{\vec{\mu}}_n)
\!\!+ \!\!\frac{1}{2}\lambda \vec{w}^{\trsp}\!\vec{w}\\
&\textbf{s.t.} \,\,\, \vec{g}_{n,f}^{\trsp}\vec{\eta}(t_n) \geq \!c_{n,f}, \!\forall f \in\{\!1,\!2,\ldots,\!F\!\}, \forall n\in\{\!1,\!2,\ldots,\!N\!\}.
\end{aligned}
\label{equ:transformed:problem}
\end{equation}
Here, the regularized term $\frac{1}{2}\lambda \vec{w}^{\trsp}\vec{w}$ with $\lambda>0$ is added into (\ref{equ:transformed:problem}) so as to alleviate the over-fitting issue\footnote{This treatment has been widely adopted in many regressions, e.g., kernel ridge regression \cite{saunders,murphy}.  }. 

We can solve (\ref{equ:transformed:problem}) by introducing Lagrange
multipliers $\alpha_{n,f}\geq 0$, leading to the Lagrange function 
\begin{equation}
\begin{aligned}
&L(\vec{w},\vec{\alpha})= \sum_{n=1}^{N}\frac{1}{2}(\vec{\Theta}^{\trsp}(t_n)\vec{w}\!-\!\hat{\vec{\mu}}_n)^{\trsp} \hat{\vec{\Sigma}}_n^{-1}
\!(\vec{\Theta}^{\trsp}(t_n)\vec{w}\!-\!\hat{\vec{\mu}}_n)\\
&+ \frac{1}{2}\lambda \vec{w}^{\trsp}\vec{w}
- \sum_{n=1}^{N}\sum_{f=1}^{F}\alpha_{n,f} (\vec{g}_{n,f}^{\trsp}\vec{\Theta}(t_n)^{\trsp}\vec{w}-c_{n,f})
\end{aligned}
\label{equ:lagrange}
\end{equation}
with $\vec{\alpha}=[\alpha_{1,1},\alpha_{1,2},\ldots,\alpha_{1,F} \ \ldots \ldots\ \alpha_{N,1},\alpha_{N,2},\ldots,\alpha_{N,F}]$.
By calculating the derivative $\frac{\partial L(\vec{w},\vec{\alpha})}{\partial\vec{w}}$ and setting as 0, we have\footnote{\emph{Woodbury identity} is used \cite{petersen2008matrix}: if $\vec{P}\succ0$ and $\vec{R}\succ0$, $(\vec{P}^{-1} + \vec{B}^{\trsp}\vec{R}^{-1}\vec{B})^{-1}\vec{B}^{\trsp}\vec{R}^{-1} = \vec{P}\vec{B}^{\trsp} (\vec{B}\vec{P}\vec{B}^{\trsp} +\vec{R})^{-1}.$}

\begin{equation}
\begin{aligned}
\vec{w}^{*}=&(\vec{\Phi}\vec{\Sigma}^{-1}\vec{\Phi}^{\trsp}+\lambda \vec{I})^{-1}(\vec{\Phi}\vec{\Sigma}^{-1}\vec{\mu}+\vec{\Phi}\overline{\vec{G}}\vec{\alpha})\\
=&
\vec{\Phi}( \vec{\Phi}^{\trsp} \vec{\Phi} +\lambda \vec{\Sigma} )^{-1} (\vec{\mu}+\vec{\Sigma}\overline{\vec{G}}\vec{\alpha}),
\end{aligned}
\label{equ:optimal:w}
\end{equation}
where 
\begin{equation}
\begin{aligned}
&\vec{\Phi}=[
\vec{\Theta}(t_1) \ \vec{\Theta}(t_2) \ \cdots \ \vec{\Theta}(t_N)
],\\
&\vec{\Sigma}=\mathrm{blockdiag}(\hat{\vec{\Sigma}}_1, \ \hat{\vec{\Sigma}}_2, \ \ldots, \ \hat{\vec{\Sigma}}_N), \quad \\
&{\vec{\mu}}=[
\hat{\vec{\mu}}_1^{\trsp} \ \hat{\vec{\mu}}_2^{\trsp} \ \cdots \ \hat{\vec{\mu}}_N^{\trsp}
]^{\trsp}, \\
&\vec{G}_n=[\vec{g}_{n,1} \ \vec{g}_{n,2} \ \ldots \ \vec{g}_{n,F}], \forall n\in\{1,\!2,\ldots,\!N\},\\
&\overline{\vec{G}}=\mathrm{blockdiag}(\vec{G}_1,\ \vec{G}_2, \ \ldots, \ \vec{G}_N). \\
\end{aligned}
\label{equ:notations:define}
\end{equation}

Furthermore, substituting the optimal $\vec{w}^{*}$ into (\ref{equ:lagrange}) gives
\begin{equation}
\widetilde{L}(\vec{\alpha})\!= \!\vec{\alpha}^{\trsp} \overline{\vec{G}}^{\trsp}\vec{\Sigma}\mathcal{A}\vec{\Sigma}\overline{\vec{G}} \vec{\alpha}+ ( 2\vec{\mu}^{\trsp} \mathcal{A} \vec{\Sigma}\overline{\vec{G}} + \overline{\vec{C}}^{\trsp} ) \vec{\alpha}+\emph{const}
\label{equ:La}
\end{equation}
with 
\begin{equation}
\begin{aligned}
&\vec{C}_n=[c_{n,1} \ c_{n,2} \ \ldots \ c_{n,F}]^{\trsp}, \forall n\in\{1,\!2,\ldots,\!N\},\\
&\overline{\vec{C}}=[\vec{C}_1^\trsp \ \vec{C}_2^\trsp \ldots \vec{C}_N^\trsp]^{\trsp},\\
&\mathcal{A}=-\frac{1}{2}(\vec{\Phi}^{\trsp}\vec{\Phi}+\lambda\vec{\Sigma})^{-1} (\vec{\Phi}^{\trsp}\vec{\Phi}\vec{\Sigma}^{-1} \vec{\Phi}^{\trsp}\vec{\Phi}+\lambda\vec{\Phi}^{\trsp}\vec{\Phi})  \\
&\quad \quad\quad\quad\quad\quad \quad\quad\quad \quad\quad(\vec{\Phi}^{\trsp}\vec{\Phi}+\lambda\vec{\Sigma})^{-1}.
\end{aligned}
\label{equ:A:ini}
\end{equation}

Similarly to KMP, we employ the well-known kernel trick $\vec{\varphi}(t_i)^{\trsp} \vec{\varphi}(t_j)=k(t_i,t_j)$, with $k(\cdot,\cdot)$ being a kernel function, to (\ref{equ:basis:function:time}), giving  

\begin{equation}
\vec{k}(t_i,t_j)\!=\! \vec{\Theta}({t_i})^{\trsp}\vec{\Theta}({t_j})\!\!=\!\!
\left[ \begin{matrix} k_{tt}(i,j)\vec{I}_{\mathcal{O}} \!&\! k_{td}(i,j)\vec{I}_{\mathcal{O}}\\
k_{dt}(i,j)\vec{I}_{\mathcal{O}} \!&\! k_{dd}(i,j)\vec{I}_{\mathcal{O}} \\
\end{matrix} \right],
\label{equ:kernel:matrix:time}
\end{equation}
where\footnote{In order to facilitate the kernel application, we approximate $\dot{\vec{\varphi}}(t)=\frac{\vec{\varphi}(t+\delta)-\vec{\varphi}(t)}{\delta}$, with a small constant $\delta>0$.} 
$
k_{tt}(i,j)=k(t_i,t_j),
k_{td}(i,j)=\frac{k(t_i,t_j+\delta)-k(t_i,t_j)}{\delta},
k_{dt}(i,j)=\frac{k(t_i+\delta,t_j)-k(t_i,t_j)}{\delta}, 
k_{dd}(i,j)=\frac{k(t_i\!+\delta, t_j+\delta) -k(t_i+\delta, t_j) -k(t_i,t_j+\!\delta) +k(t_i,t_j)}{{\delta}^{2}}. 
$
Subsequently, combined with the definition of $\vec{\Phi}$ in (\ref{equ:notations:define}), we have the kernel matrix
\begin{equation}
\vec{K}
=\vec{\Phi}^{\trsp}\vec{\Phi}=\left[\begin{matrix} 
\vec{k}(t_1, t_1) & \vec{k}(t_1, t_2) & \cdots &\vec{k}(t_1, t_N) \\
\vec{k}(t_2, t_1) & \vec{k}(t_2, t_2) & \cdots &\vec{k}(t_2, t_N) \\
\vdots & \vdots &  \ddots & \vdots \\
\vec{k}(t_N, t_1) & \vec{k}(t_N, t_2) & \cdots &\vec{k}(t_N, t_N) \\
\end{matrix}\right].
\label{equ:K:matrix}
\end{equation}
Thus, we can rewrite $\mathcal{A}$ in (\ref{equ:A:ini}) as
\begin{equation}
\begin{aligned}
\mathcal{A}=-\frac{1}{2}(\vec{K}+\lambda\vec{\Sigma})^{-1} (\vec{K}\vec{\Sigma}^{-1} \vec{K}+\lambda \vec{K}) 
(\vec{K}+\lambda\vec{\Sigma})^{-1}
\end{aligned}
\label{equ:A}
\end{equation}

Let us now revisit the function $\widetilde{L}(\vec{\alpha})$ of Lagrange multipliers in (\ref{equ:La}), and denote
\begin{equation}
\begin{aligned}
\mathcal{B}_1&=\overline{\vec{G}}^{\trsp}\vec{\Sigma}\mathcal{A}\vec{\Sigma}\overline{\vec{G}},\\
\mathcal{B}_2&=2\vec{\mu}^{\trsp} \mathcal{A} \vec{\Sigma}\overline{\vec{G}} + \overline{\vec{C}}^{\trsp}.
\end{aligned}
\label{equ:qp:coef}
\end{equation}
Thus, we can tackle the problem of finding optimal Lagrange multipliers $\vec{\alpha}$ through maximizing
\begin{equation}
\begin{aligned}
\widetilde{L}(\vec{\alpha})= &\vec{\alpha}^{\trsp}\mathcal{B}_1 \vec{\alpha}+ \mathcal{B}_2 \vec{\alpha}, \\
s.t. \ &\vec{\alpha}\geq \vec{0}.
\end{aligned}
\label{equ:qp}
\end{equation}
It is noted that ${\mathcal{A}=\mathcal{A}^{\trsp}\preceq 0}$ and hence ${\mathcal{B}_1=\mathcal{B}_1^{\trsp}\preceq 0}$. So, the problem described in (\ref{equ:qp}) is a typical quadratic optimization problem with linear constraints, which can be solved by the classical \emph{quadratic programming}. 

Once the optimal $\vec{\alpha}^{*}$ is determined, we can apply (\ref{equ:optimal:w}) to the prediction problem. Namely, given a query input $t^{*}$, we have its corresponding output as
\begin{equation}
\begin{aligned}
\vec{\eta}(t^{*})&=\left[\begin{array}{c}
\vec{\xi}(t^{*}) \\ \dot{\vec{\xi}}(t^{*})
\end{array}\right]
=\vec{\Theta}^{\trsp}(t^{*})\vec{w}^{*} \\
&=\vec{\Theta}^{\trsp}(t^{*})\vec{\Phi}( \vec{\Phi}^{\trsp} \vec{\Phi} +\lambda \vec{\Sigma} )^{-1} (\vec{\mu}+\vec{\Sigma}\overline{\vec{G}}\vec{\alpha}^{*})\\
&=\vec{k}^{*}( \vec{K} +\lambda \vec{\Sigma} )^{-1} (\vec{\mu}+\vec{\Sigma}\overline{\vec{G}}\vec{\alpha}^{*}),
\end{aligned}
\label{equ:prediction}
\end{equation}
where 
\begin{equation}
\vec{k}^{*}=\left[ 
\vec{k}(t^{*}, t_1) \ \vec{k}(t^{*}, t_2) \ \cdots \ \vec{k}(t^{*}, t_N)\right].
\label{equ:kernel:matrix:pre}
\end{equation}
Until now, we have explained the linearly constrained imitation learning framework, which we refer to as \emph{linearly constrained KMP} (LC-KMP).
A complete algorithm is summarized in Algorithm~\ref{alg:lc-kmp}. Before ending this section, we show a few insights over LC-KMP: 
\begin{enumerate}
\item \textbf{Non-constrained Learning}: if we consider imitation learning without linear constraints, i.e., $\vec{\alpha}=0$, the prediction of LC-KMP in (\ref{equ:prediction}) will become the vanilla KMP.
\item \textbf{Partially-constrained Learning:} when only partial linear constraints are required, we can deactivate the rest constraints by simply setting their corresponding Lagrange multipliers as zero, which can be ensured by adding an additional equality constraint over $\vec{\alpha}$. 
\item \textbf{Equality-constrained Learning:} the framework in (\ref{equ:ini:problem}) can be used to tackle equality constraints.
For example, the equality constraint $\vec{g}_{n,f}^{\trsp}\vec{\eta}(t_n) = c_{n,f}$ can be the guaranteed by $\vec{g}_{n,f}^{\trsp}\vec{\eta}(t_n) > c_{n,f}-\epsilon$ and $-\vec{g}_{n,f}^{\trsp}\vec{\eta}(t_n) > -c_{n,f}-\epsilon$, with an approximation error $\epsilon>0$. 
\item \textbf{Adaptations with constraints:} we can extend LC-KMP to adapt trajectories towards arbitrary desired points in terms of position and velocity\footnote{The adaptation property has been proven essential in many applications, such as grasping an object at difference locations \cite{stulp2011learning} or striking a ping-pong ball at a desired position while having a desired velocity \cite{huang2016}.}. 
Given $L$ desired points $\overline{\vec{D}}=\{\bar{t}_l,\bar{\vec{\mu}}_l, \bar{\vec{\Sigma}}_l\}_{l=1}^{L}$ comprising the desired time $\bar{t}_l$ and its corresponding output distribution (i.e., 
$\overline{\vec{\eta}}(\bar{t}_l) \sim \mathcal{N}(\bar{\vec{\mu}}_l,\bar{\vec{\Sigma}}_l)$), we can directly concatenate\footnote{Please refer to \cite{huang2017} for the details of updating the reference trajectory.} 
$\overline{\vec{D}}$ and ${\vec{D}}$ to obtain an \emph{extended probabilistic reference trajectory} ${\vec{D}}^{U}$. After that, we exploit ${\vec{D}}^{U}$ instead of $\vec{D}$ in the framework (\ref{equ:ini:problem}), which will generate a trajectory that passes through various desired points while satisfying additional linear constraints. 

\end{enumerate}

\begin{algorithm}[bt]
	\caption{\emph{Linearly Constrained KMP}}
	\begin{algorithmic}[1]
		\State \textbf{Initialization}
		\Statex - Set $\lambda$ and $k(\cdot,\cdot)$
		\Statex - Define linear constraints $\{\{\vec{g}_{n,f},\vec{c}_{n,f}\}_{n=1}^{N}\}_{f=1}^{F}$	
		\Statex - Collect demonstrations $\{\{t_{n,m},\vec{\xi}_{n,m},\dot{\vec{\xi}}_{n,m}\}_{n=1}^{N}\}_{m=1}^{M}$
		\State \textbf{Model demonstrations} (Section~\ref{sec:gmm})
		\Statex - Model the joint probability distribution $\mathcal{P}(t,\vec{\xi},\dot{\vec{\xi}})$\
		\Statex - Retrieve the reference trajectory $\vec{D}_r\!=\!\{t_n,\!\hat{\vec{\mu}}_n,\!\hat{\vec{\Sigma}}_n\}_{n=1}^{N}$ 		
		\State \textbf{Optimize Lagrange multipliers} (Section~\ref{sec:lc-kmp})
		\Statex - Calculate $\vec{\mu}$, $\vec{\Sigma}$, $\overline{\vec{G}}$, $\overline{\vec{C}}$, $\vec{K}$ via (\ref{equ:notations:define}), (\ref{equ:A:ini}), (\ref{equ:kernel:matrix:time}), (\ref{equ:K:matrix})
		\Statex - Calculate $\mathcal{A}$, $\mathcal{B}_1$, $\mathcal{B}_2$ via (\ref{equ:A}) and (\ref{equ:qp:coef}).
		\Statex - Find optimal $\vec{\alpha}^{*}$ by solving (\ref{equ:qp})		
		\State \textbf{Prediction} (Section~\ref{sec:lc-kmp})
		\Statex - For a given input $t^{*}$, compute $\vec{k}^{*}$ using (\ref{equ:kernel:matrix:pre})
		\Statex - Predict $\vec{\eta}(t^{*})$ using (\ref{equ:prediction})
	\end{algorithmic}
	\label{alg:lc-kmp}
\end{algorithm}

\begin{table*}[bt]
	\caption {Comparison Between MPC and LC-KMP}
	\vspace{-0.2cm}
	\centering
	\scalebox{0.88}{	
		\begin{tabular}{lccc}
			\toprule %
			&\textbf{MPC} & \textbf{LC-KMP}\\ \toprule %
			$\textbf{Model}$ 
			& $\vec{\eta}({t+1})=
			{\vec{A}}
			\vec{\eta}({t})+
			{\vec{B}}{\vec{u}}(t)$
			& $\vec{\eta}(t)=
			\vec{\Theta}^{\trsp}(t)\vec{w}$
			\\
			\midrule
			$\textbf{Referecne \ traj.}$
			& $\{\hat{\vec{\eta}}_t\}_{t=1}^{N}$
			& $\{\hat{\vec{\mu}}_n,\hat{\vec{\Sigma}}_n\}_{n=1}^{N}$ 
			\\
			\midrule
			$\textbf{Optimization}$
			&  $\begin{aligned}
			\argmin_{\vec{u}_1,\ldots,\vec{u}_{N-1}}& \sum_{t=1}^{N}(\vec{\eta}_t-\hat{\vec{\eta}}_t)^{\trsp}\!\vec{Q}_t(\vec{\eta}_t-\hat{\vec{\eta}}_t)+\sum_{t=1}^{N-1}{\vec{u}}_t^{\trsp}\vec{R}_t{\vec{u}}_t\\
			\textbf{s.t.} \quad & \vec{\eta}_{max} \geq \vec{\eta}_t \geq \vec{\eta}_{min}, \ \forall t\in\{2,3,\ldots,N\}\\
			&\vec{u}_{max} \geq \vec{u}_t \geq \vec{u}_{min}, \ \forall t\in\{1,2,\ldots,N-1\}\\
			\end{aligned}$
			&$
			\begin{aligned}
			&\argmin_{\vec{w}} \!\! \sum_{n=1}^{N}\!\!\frac{1}{2}\!(\vec{\Theta}^{\trsp}(t_n)\vec{w}\!-\!\hat{\vec{\mu}}_n)^{\trsp}\! \hat{\vec{\Sigma}}_n^{-1}\!
			\!(\vec{\Theta}^{\trsp}(t_n)\vec{w}\!-\!\hat{\vec{\mu}}_n)
			\!\!+ \!\!\frac{1}{2}\lambda \vec{w}^{\trsp}\!\vec{w}\\
			&\textbf{s.t.} \,\,\, \vec{g}_{n,f}^{\trsp}\vec{\eta}(t_n) \geq \!c_{n,f}, \!\forall f \in\{1,2,\ldots,F\}, \forall n\in\{1,2,\ldots,N\}.
			\end{aligned}
			$
			\\
			\midrule
			$\textbf{Compact \ form}$
			& $\begin{aligned}
			\argmin_{\overline{\vec{u}}}& (\mathcal{S}_{u}\overline{\vec{u}}+\mathcal{S}_{\eta}\vec{\eta}_1-\widehat{\vec{\eta}})^{\trsp}\overline{\vec{Q}}(\mathcal{S}_{u}\overline{\vec{u}}+\mathcal{S}_{\eta}\vec{\eta}_1-\widehat{\vec{\eta}})+
			\overline{\vec{u}}^{\trsp}\overline{\vec{R}}\overline{\vec{u}}\\
			&\textbf{s.t.} \,\,\, {\vec{W}_1}\overline{\vec{u}}  \geq {\vec{W}_2}+{\vec{V}}\vec{\eta}_1,
			\end{aligned}$
			& $\begin{aligned}
			\argmin_{\vec{w}}& \frac{1}{2}(\vec{\Phi}^{\trsp}\vec{w}-{\vec{\mu}})^{\trsp} \vec{\Sigma}^{-1}
			(\vec{\Phi}^{\trsp}\vec{w}-{\vec{\mu}})
			\!+ \!\lambda \vec{w}^{\trsp}\!\vec{w}\\
			&\textbf{s.t.}  \,\,\, \overline{\vec{G}}^{\trsp}\vec{\Phi}^{\trsp}\vec{w} \geq \overline{\vec{C}}.
			\end{aligned}$
			\\
			\midrule
			$\textbf{Solution}$
			& $\overline{\vec{u}}^{*}=[\vec{u}_1^{{*}\trsp} \ \vec{u}_2^{{*}\trsp} \ \ldots \vec{u}_{N-1}^{{*}\trsp}]^{\trsp}$, apply $\vec{u}_1^{{*}}$ as control command 
			& $\vec{\alpha}^{*}$ and $\vec{w}^{*}$, apply $\vec{k}^{*}( \vec{K} +\lambda \vec{\Sigma} )^{-1} (\vec{\mu}+\vec{\Sigma}\overline{\vec{G}}\vec{\alpha}^{*})$ to trajectory prediction.
			\\
			\bottomrule
	\end{tabular}}
	\label{table:comp:table}
	\vspace{-0.5cm}
\end{table*}

\section{Connection with Model Predictive Control \label{sec:mpc}}
In this section, we provide a connection between LC-KMP and linear MPC. 
For more details about MPC, see, e.g., \cite{alessio2009survey,rawlings2000tutorial}.
Assuming that we have a linear system, described by\footnote{Note that $\vec{\eta}_t$ is interchangeably used with $\vec{\eta}(t)$. }
\begin{equation}
\vec{\eta}_{t+1}=
{\vec{A}}
\vec{\eta}_{t}+
{\vec{B}}{\vec{u}}_t,
\label{equ:sys:linear}
\end{equation}
with $\vec{A}\in \mathbb{R}^{2\mathcal{O}\times2\mathcal{O}}$ and $\vec{B}\in \mathbb{R}^{2\mathcal{O}\times \mathcal{O}}$ being the state and control matrices, and $\vec{u}_t \in \mathbb{R}^{\mathcal{O}}$ denoting the control command. The optimization objective in MPC is
\begin{equation}
\begin{aligned}
\argmin_{\vec{u}_1,\ldots,\vec{u}_{N-1}}& \sum_{t=1}^{N}(\vec{\eta}_t-\hat{\vec{\eta}}_t)^{\trsp}\!\vec{Q}_t(\vec{\eta}_t-\hat{\vec{\eta}}_t)+\sum_{t=1}^{N-1}{\vec{u}}_t^{\trsp}\vec{R}_t{\vec{u}}_t\\
\textbf{s.t.} \quad & \vec{\eta}_{max} \geq \vec{\eta}_t \geq \vec{\eta}_{min}, \ \forall t\in\{2,3,\ldots,N\}\\
&\vec{u}_{max}  \geq \vec{u}_t \geq \vec{u}_{min} , \ \forall t\in\{1,2,\ldots,N-1\}\\
\end{aligned}
\label{equ:mpc:problem}
\end{equation}
with $\vec{Q}\succeq $ and $\vec{R}\succ 0$. $\hat{\vec{\eta}}_t$ denotes the desired reference trajectory point at time $t$, while $\vec{\eta}_{min}$, $\vec{\eta}_{max}$, $\vec{u}_{min}$ and $\vec{u}_{max}$ represent the predefined limits.

Following the prediction strategy in MPC (e.g., \cite{calinon2016}), we can obtain
\begin{equation}
\underbrace{\left[\begin{matrix}
	\vec{\eta}_{1} \\\vec{\eta}_{2} \\ \vec{\eta}_{3} \\  \vdots \\ \vec{\eta}_{N}
	\end{matrix}\right]}_{\overline{\vec{\eta}}}\!\!\!=\!\!\!
\underbrace{\left[\begin{matrix}
	\vec{I} \\\!\!\!\vec{A} \\ \vec{A}^2 \\  \vdots \\ \vec{A}^{N-1}
	\end{matrix}\right]}_{\mathcal{S}_{\eta}} \!\! \vec{\eta}_1 \!\!+\!\!
\underbrace{\left[\begin{matrix}
	\vec{0} \!&\! \vec{0} \!&\! \cdots \!&\! \vec{0} \\ \vec{B} \!&\! \vec{0} \!&\! \cdots \!&\! \vec{0} \\ \vec{AB} \!&\! \vec{B} \!&\! \cdots \!&\! \vec{0} \\
	\vdots \!&\! \vdots \!&\! \ddots \!&\! \vec{0} \\
	\vec{A}^{N-2}\vec{B} \!&\! \vec{A}^{N-3}\vec{B} \!&\! \cdots \!&\! \vec{B} 
	\end{matrix}\right]}_{\mathcal{S}_{u}} \!\!\!
\underbrace{\left[\begin{matrix}
	{\vec{u}}_1 \\ {\vec{u}}_{2} \\  \vdots \\ {\vec{u}}_{N-1}
	\end{matrix}\right]}_{\overline{\vec{u}}}\!\!.
\label{equ:pred:mpc}
\end{equation}

Using (\ref{equ:pred:mpc}), we can reformulate (\ref{equ:mpc:problem}) as
\begin{equation}
\begin{aligned}
\argmin_{\overline{\vec{u}}}& (\overline{\vec{\eta}}-\widehat{\vec{\eta}})^{\trsp}\overline{\vec{Q}}(\overline{\vec{\eta}}-\widehat{\vec{\eta}})+
\overline{\vec{u}}^{\trsp}\overline{\vec{R}}\overline{\vec{u}}\\
&\textbf{s.t.} \,\,\, {\vec{W}_1}\overline{\vec{u}} \geq {\vec{W}_2}+{\vec{V}}\vec{\eta}_1,
\end{aligned}
\label{equ:mpc:problem:compact}
\end{equation}
where 
\begin{equation}
\begin{aligned}
\overline{\vec{\eta}}&=[\vec{\eta}_1^{\trsp} \ \vec{\eta}_2^{\trsp} \ \ldots \vec{\eta}_N^{\trsp}]^{\trsp},\\
\widehat{\vec{\eta}}&=[\hat{\vec{\eta}}_1^{\trsp} \ \hat{\vec{\eta}}_2^{\trsp} \ \ldots \hat{\vec{\eta}}_N^{\trsp}]^{\trsp},\\
\overline{\vec{u}}&=[\vec{u}_1^{\trsp} \ \vec{u}_2^{\trsp} \ \ldots \vec{u}_{N-1}^{\trsp}]^{\trsp},\\
\overline{\vec{Q}}&=\mathrm{blockdiag}(\vec{Q}_1, \ \vec{Q}_2, \ \ldots, \ \vec{Q}_N),\\
\overline{\vec{R}}&=\mathrm{blockdiag}(\vec{R}_1, \ \vec{R}_2, \ \ldots, \ \vec{R}_{N-1}).\\
\end{aligned}
\end{equation}
Note that $\vec{W}_1$, $\vec{W}_2$ and $\vec{V}$ denote coefficient matrices, which can be determined by using $\mathcal{S}_{\eta}$, $\mathcal{S}_{u}$, $\vec{\eta}_{min}$, $\vec{\eta}_{max}$, $\vec{u}_{min}$ and $\vec{u}_{max}$.

Furthermore, 
substituting $\overline{\vec{\eta}}=\mathcal{S}_{u}\overline{\vec{u}}+\mathcal{S}_{\eta}\vec{\eta}_1$ into (\ref{equ:mpc:problem:compact}) gives
\begin{equation}
\begin{aligned}
\argmin_{\overline{\vec{u}}}& (\mathcal{S}_{u}\overline{\vec{u}}+\mathcal{S}_{\eta}\vec{\eta}_1-\widehat{\vec{\eta}})^{\trsp}\overline{\vec{Q}}(\mathcal{S}_{u}\overline{\vec{u}}+\mathcal{S}_{\eta}\vec{\eta}_1-\widehat{\vec{\eta}})+
\overline{\vec{u}}^{\trsp}\overline{\vec{R}}\overline{\vec{u}}\\
&\textbf{s.t.} \,\,\, {\vec{W}_1}\overline{\vec{u}}  \geq {\vec{W}_2}+{\vec{V}}\vec{\eta}_1.
\end{aligned}
\label{equ:mpc:compact:new}
\end{equation}

Now, let us recall the optimization problem (\ref{equ:transformed:problem}) of LC-KMP, whose impact form is
\begin{equation}
\begin{aligned}
\argmin_{\vec{w}}& \frac{1}{2}(\vec{\Phi}^{\trsp}\vec{w}-{\vec{\mu}})^{\trsp} \vec{\Sigma}^{-1}
(\vec{\Phi}^{\trsp}\vec{w}-{\vec{\mu}})
\!+ \!\frac{1}{2}\lambda \vec{w}^{\trsp}\!\vec{w}\\
&\textbf{s.t.}  \,\,\, \overline{\vec{G}}^{\trsp}\vec{\Phi}^{\trsp}\vec{w} \geq \overline{\vec{C}}.
\end{aligned}
\label{equ:lc:kmp:compact}
\end{equation}

By comparing (\ref{equ:mpc:compact:new}) and (\ref{equ:lc:kmp:compact}), it can be seen that both optimization problems will be equivalent\footnote{The scalar ``$\frac{1}{2}$" can be ignored as it does not influence the optimization.} \emph{if} 
$\mathcal{S}_{u}=\vec{\Phi}^{\trsp}$, $\widehat{\vec{\eta}}={\vec{\mu}}+\mathcal{S}_{\eta}\vec{\eta}_1$, $\overline{\vec{Q}}=\vec{\Sigma}^{-1}$, $\overline{\vec{R}}=\lambda\vec{I}$, $\vec{W}_1=\overline{\vec{G}}^{\trsp}\vec{\Phi}^{\trsp}$ and $\vec{W}_2=\overline{\vec{C}}-{\vec{V}}\vec{\eta}_1$. 
In fact, the key differences between MPC and LC-KMP lie at two aspects:
\begin{enumerate}
	\item In contract to MPC that uses the dynamics model (\ref{equ:sys:linear}) for predicting future trajectory, LC-KMP uses the parametric model (\ref{equ:para:traj}) instead. Note that basis functions in LC-KMP can be ultimately alleviated through the kernel trick, resulting in a non-parametric approach. 
	\item MPC aims at finding the optimal control command $\overline{\vec{u}} \in \mathbb{R}^{(N-1)\mathcal{O}}$, while LC-KMP aims for the optimal trajectory parameter $\vec{w} \in \mathbb{R}^{B\mathcal{O}}$.
\end{enumerate} 
For the purpose of clear comparison, we summarize the differences between MPC and LC-KMP in Table~\ref{table:comp:table}.

It is worth pointing out that the unconstrained MPC was studied in \cite{calinon2014task,calinon2016}, where setting $\hat{\vec{\eta}}_t=\hat{\vec{\mu}}_t$ and $\vec{Q}_t=\hat{\vec{\Sigma}}_t^{-1}$ in (\ref{equ:mpc:problem}) yields the \emph{minimal intervention control} problem
\begin{equation}
\begin{aligned}
\argmin_{\vec{u}_1,\ldots,\vec{u}_{N-1}}& \sum_{t=1}^{N}(\vec{\eta}_t-\hat{\vec{\mu}}_t)^{\trsp}\!\hat{\vec{\Sigma}}_t^{-1}(\vec{\eta}_t-\hat{\vec{\mu}}_t)+\sum_{t=1}^{N-1}{\vec{u}}_t^{\trsp}\vec{R}_t{\vec{u}}_t,
\end{aligned}
\label{equ:min:intervention:problem}
\end{equation}
which is equivalent to
\begin{equation}
\begin{aligned}
\argmin_{\overline{\vec{u}}} (\mathcal{S}_{u}\overline{\vec{u}}\!+\!\mathcal{S}_{\eta}\vec{\eta}_1\!-\!\vec{\mu})^{\trsp}{\vec{\Sigma}}^{-1}(\mathcal{S}_{u}\overline{\vec{u}}\!+\!\mathcal{S}_{\eta}\vec{\eta}_1\!-\!\vec{\mu})+
\overline{\vec{u}}^{\trsp}\overline{\vec{R}}\overline{\vec{u}}.
\end{aligned}
\label{equ:min:intervention:problem:compact}
\end{equation}
We can find that
minimal intervention control, as a special case of unconstrained MPC, shares similarities with LC-KMP in terms of \emph{imitation}, since minimal intervention control exploits the distribution of demonstrations as well.

\begin{figure*}[bt] \centering 
	\includegraphics[width=0.88\textwidth]{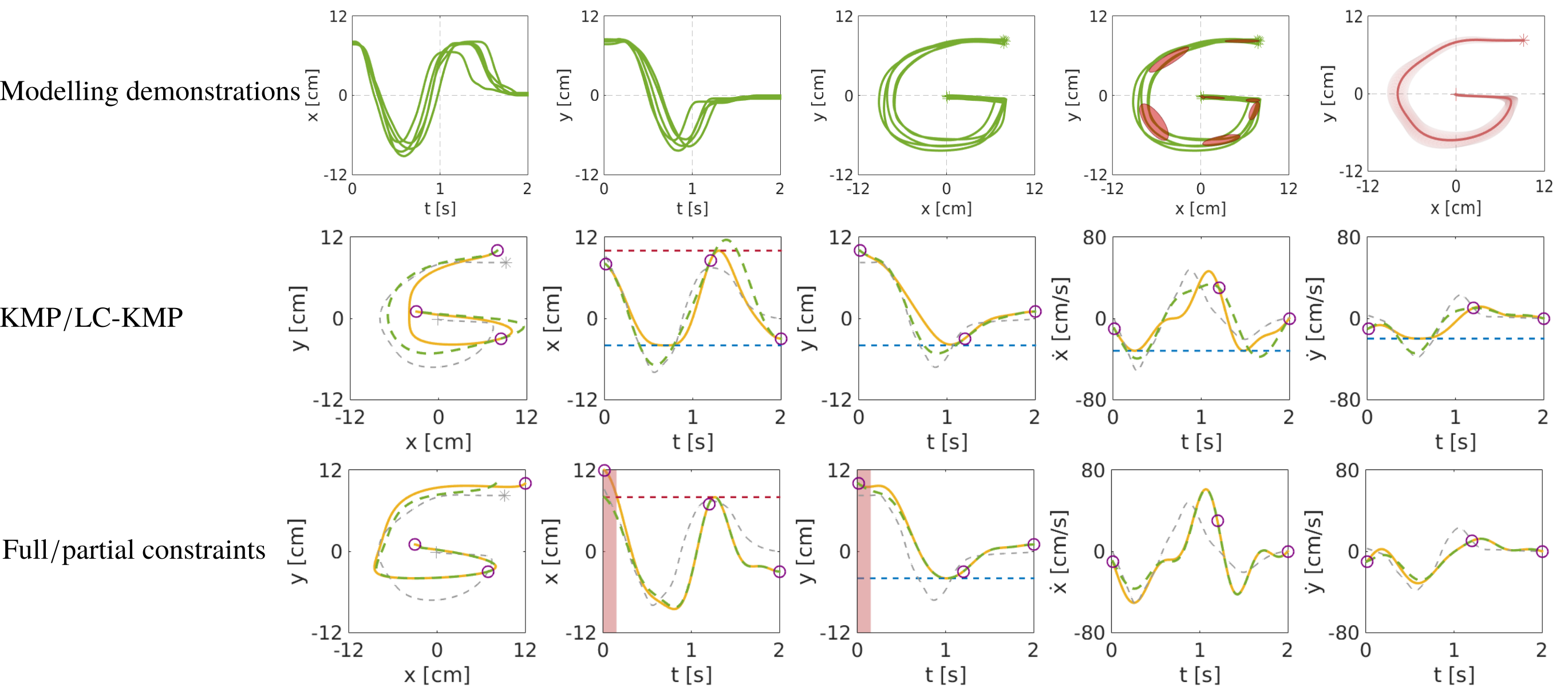}
	\vspace{-0.32cm}
	\caption{Learning and adaptation of 2D letter `G' under motions limits. \emph{Top row} shows demonstrations (green curves) and the corresponding GMM/GMR modeling results. Ellipses represent Gaussian components in GMM, while the pink curve and the shaded pink area respectively depict the mean and the standard deviation of the retrieved trajectory by GMR.  \emph{Middle row} shows adaptations by using LC-KMP (solid yellow curves) and vanilla KMP (dashed green curves), where the dashed gray curves denote the mean of the probabilistic reference trajectory. Circles represent desired points. The dashed blue and red curves correspond to the lower and upper limits, respectively. \emph{Bottom row} depicts adaptations with full constraints (solid yellow curves) and partial constraints (dashed green curves) with the shaded red area corresponding to the inactive region. }
	\label{fig:g:2d} 
\end{figure*}

\begin{figure*}[bt] \centering 
	\vspace{-0.32cm}
	\includegraphics[width=0.83\textwidth]{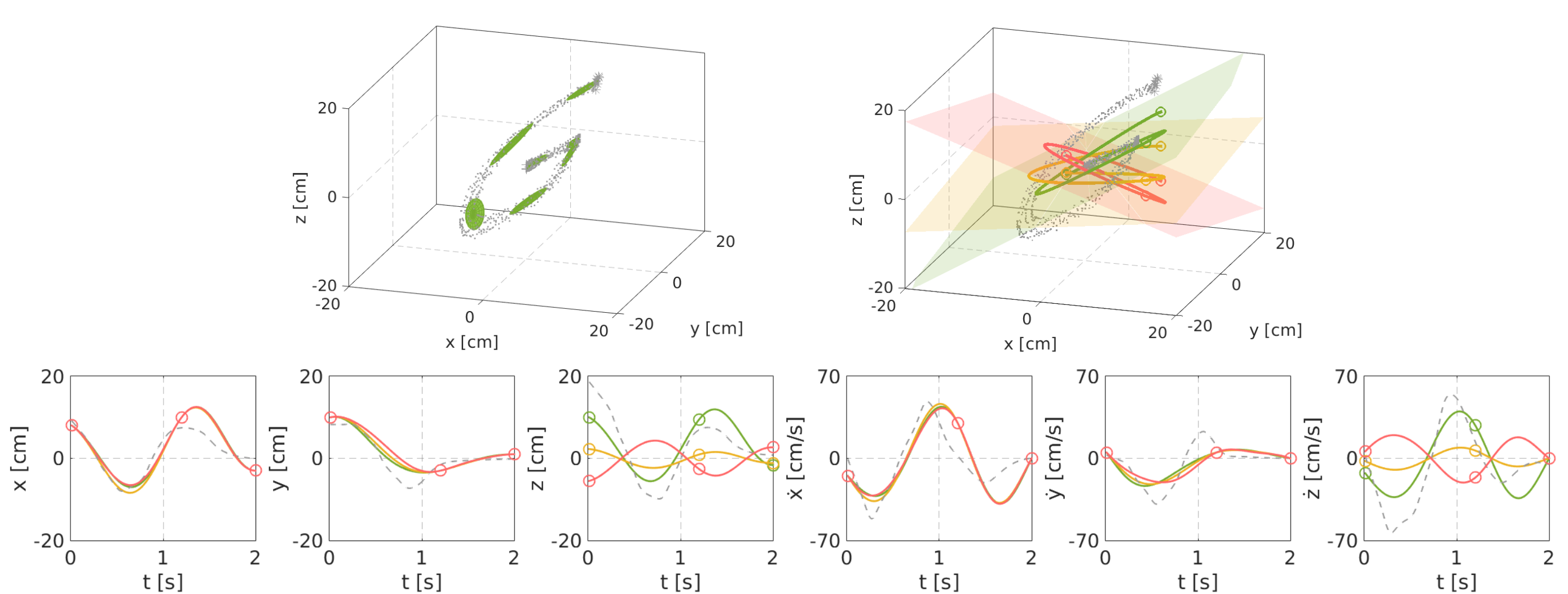}
	\vspace{-0.32cm}
	\caption{Adaptation of 3D letter `G' in different planes. \emph{Top-left} shows GMM modeling of demonstrations, while \emph{top-right} and \emph{bottom} plot trajectory adaptations in different planes (shown by colored planes), where circles denote desired points.}
	\label{fig:ada:3d:xyz:plane} 
\end{figure*}

\section{Evaluations \label{sec:eva}}

In order to verify our framework, several evaluations are provided, including
	(i) adapting 2D trajectories with/without motion limits, as well as adaptations with full/partial motion limits (Section~\ref{subsec:eva:limit});
	(ii) adapting 3D trajectories in various planes (Section~\ref{subsec:eva:plane});
    (iii) generating stable waking trajectories for a humanoid robot (Section~\ref{subsec:eva:stable}). 
The Gaussian kernel $k(t_i,t_j)=\exp(-k_h (t_i-t_j)^2)$ is used in this section.

\subsection{Adaptation with Motion Limits\label{subsec:eva:limit}}
We first apply LC-KMP to the learning and adaptation of 2D hand-written letter `G', where five demonstrations comprising input $t$ and 2D output
$\vec{\xi}(t)=[x(t) \ y(t)]^\trsp$ are used (plotted by solid green curves in Fig.~\ref{fig:g:2d} (\emph{top row})). Following Algorithm~\ref{alg:lc-kmp}, we use GMM to model the demonstrations and later use GMR to retrieve the probabilistic reference trajectory. As a comparison, we separately apply LC-KMP and vanilla KMP to adapt trajectories towards desired points under motion limits. 
We consider the following constraints: 
\begin{equation}
-4 \leq x \leq 10, \ y \geq  -4, \
\dot{x} \geq  -32, \ \dot{y} \geq  -20.\\
\end{equation}
Other relevant parameters are $\lambda=3$ and $k_h=6$.
It can be seen from Fig.~\ref{fig:g:2d} (\emph{middle row}) that LC-KMP (solid yellow curves) indeed modulates trajectories towards various desired points (depicted by circles) while respecting the motions limits, where lower and upper limits are shown by blue and red dashed curves, respectively. In contrast, vanilla KMP (dashed green curves) only focuses on trajectory adaptations, ignoring the motions limits.

Furthermore, we test LC-KMP with full constraints and partial constraints. Specifically, in the former case constraints are active over the whole time duration (i.e., $0 < t \leq 2$), while in the latter case constraints are only active when $0.15 <  t \leq 2$.  We consider the following constraints
\begin{equation}
x \leq 8 \ , \ y \geq  -4.
\end{equation}
As shown in Fig.~\ref{fig:g:2d} (\emph{bottom row}), the upper limit of $x(t)$ has a conflict with the $x$ component of the first desired point, thus the adapted trajectory (dashed green curves), that obeys the constraints over the whole time duration, fails to passing through the $x$ component of the first desired point. In contrast, when only partial constraints are active, the adapted trajectory (solid yellow curves) is capable of passing through the first desired point. Note that once the adapted trajectory moves out of the deactivate region (depicted by the shaded red area), it will comply with the constraints again.

\begin{figure*}[bt] \centering 
	\includegraphics[width=0.91\textwidth]{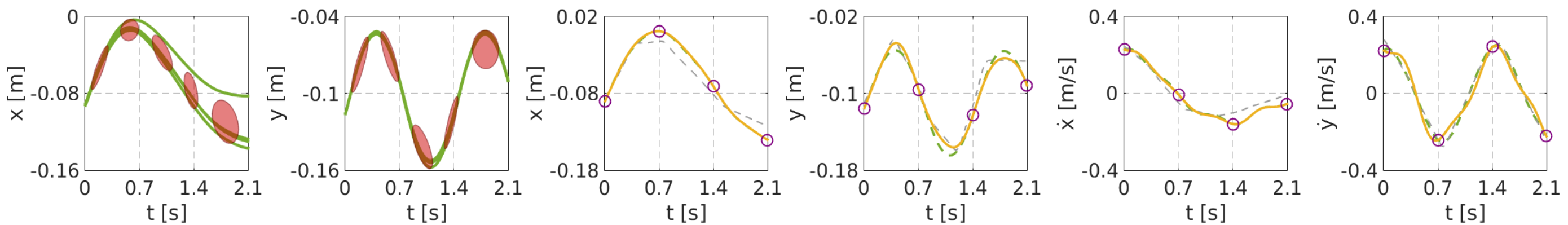}
	\vspace{-0.30cm}
	\caption{Learning and adapting walking trajectories. Green solid curves and ellipses denote demonstrations and GMM components, respectively. Yellow solid curves and green dashed curves respectively represent adapted trajectories by using LC-KMP and vanilla KMP. Circles depict desired points.}
	\label{fig:loco:ada} 
\end{figure*}

\begin{figure*}[bt] \centering 
	\includegraphics[width=0.72\textwidth]{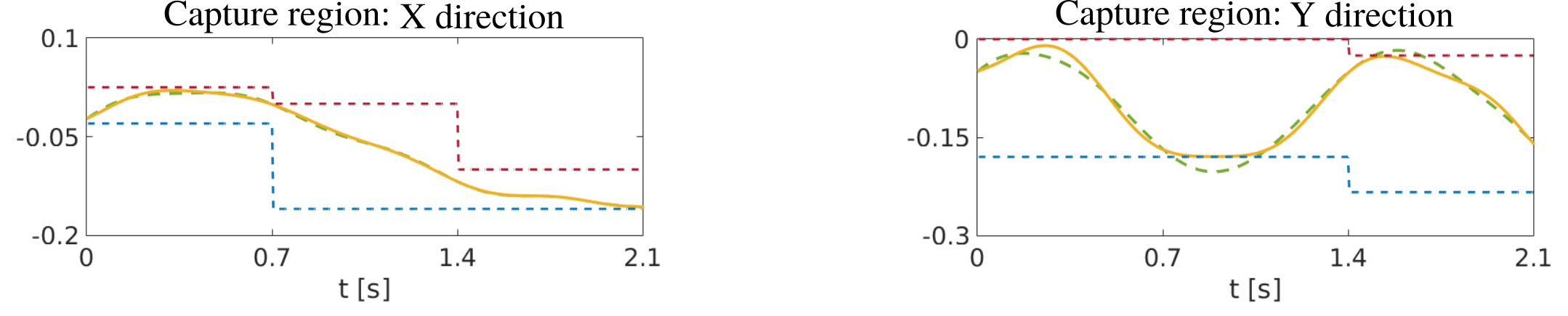}
	\vspace{-0.45cm}
	\caption{Evaluations of stability criteria, where yellow solid curves and green dashed curves correspond to LC-KMP and vanilla KMP, respectively. Dashed red and blue curves depict the upper and lower bounds of capture regions, respectively.}
	\label{fig:capture:region} 
\end{figure*}

\begin{figure*}[bt] \centering 
	\includegraphics[width=0.81\textwidth]{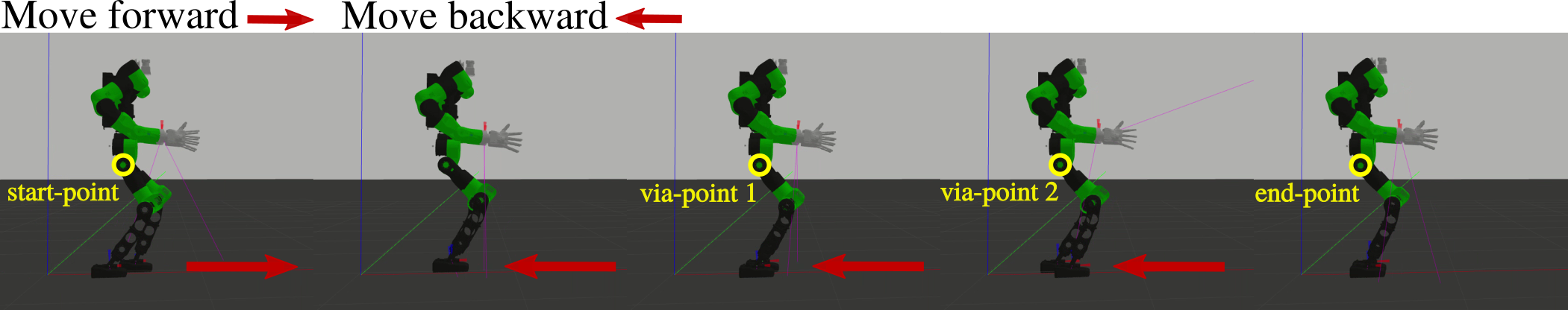}
	\vspace{-0.30cm}
	\caption{Snapshots of walking movement of a simulated humanoid robot, where  LC-KMP is used to plan the CoM trajectories.}
	\label{fig:loco:simu} 
\end{figure*}

\subsection{Adaptation with Plane Constraint \label{subsec:eva:plane}}

We here consider adaptations of 3D letter `G' in different planes. Five demonstrations (depicted by gray dots in Fig.~\ref{fig:ada:3d:xyz:plane} (\emph{top-left})) in terms of input $t$ and 3D output $\vec{\xi}(t)=[x(t) \ y(t) \ z(t)]^\trsp$ are collected. Similarly, demonstrations are modeled by GMM and the probabilistic reference trajectory is extracted via GMR, which is used to train LC-KMP. The relevant parameters in LC-KMP are $\lambda=5$ and $k_h=2$.
The plane constraint is defined as 
\begin{equation}
\begin{aligned}
a_x x + b_y y  +c_z z = d.
\end{aligned}
\end{equation}
We have three groups of evaluations, whose corresponding parameters are set as:
(\emph{i}) $a_x^{(1)}=1, b_y^{(1)}=0.2, c_z^{(1)}=-1.1, d^{(1)}=-1$;
(\emph{ii}) $a_x^{(2)}=0.6, b_y^{(2)}=0.4, c_z^{(2)}=-3, d^{(2)}=2$;
(\emph{iii}) $a_x^{(3)}=1, b_y^{(3)}=0.6, c_z^{(3)}=2, d^{(3)}=3$.
Evaluations are provided in Fig.~\ref{fig:ada:3d:xyz:plane}, showing that LC-KMP can learn and adapt demonstrations into different planes and meanwhile considering various start-/via-/end- points in terms of both position and velocity. 

\subsection{Stable Walking Trajectories for Humanoid Robot \label{subsec:eva:stable}} 

We now consider a more challenging task, where a stable walking trajectory for a humanoid robot is required. 
As suggested in \cite{pratt2006capture, koolen2012capturability}, the \emph{capture region} which is defined by proper constraints over the position and velocity profiles of the center of mass (CoM) can be used to ensure the stability. In this example, we aim to plan CoM trajectories to accomplish the non-periodic walking (i.e., switch from forward motion to backward motion) over three periods (each period lasts $0.7s$). Specifically, we design the capture regions as\footnote{The parameters of linear constraints are determined according to the physical features of the simulated humanoid platform \cite{muratore2017xbotcore}.}
\begin{equation}
\begin{aligned}
x_l& \leq a_x  x+ b_x \dot{x} \leq x_u \\
y_l& \leq a_y  y+ b_y \dot{y} \leq y_u
\end{aligned}
\end{equation}
with $a_x=a_y=1$, $b_x=b_y=(\frac{h_{com}}{g})^{\frac{1}{2}}$, where $h_{com}=0.8898$ and $g = 9.8$. $x_l$, $x_u$, $y_l$ and $y_u$ are set by
(\emph{i})
 $\mathrm{if} \ \!0 \!< \!t\! \leq \!0.7, x_l\!=\!-0.03, x_u\!=\!0.025, y_l\!=\!-0.18, y_u\!=\!0$;
(\emph{ii}) $\mathrm{if} \ 0.7 \!< \!t \!\leq \!1.4, x_l\!=\!-0.16, x_u\!=\!0, y_l\!=\!-0.18, y_u\!=\!0$;
(\emph{iii}) $\mathrm{if} \ 1.4 \!< \!t\! \leq \!2.1, x_l\!=\!-0.16, x_u\!=\!-0.1, y_l\!=\!-0.234, y_u\!=\!-0.025$. LC-KMP parameters are $\lambda=6$ and $k_h=2$.

We first use optimization solver \cite{ding2019nonlinear} to generate four training trajectories (solid green curves in Fig.~\ref{fig:loco:ada}), which serve as demonstrations for our framework. 
For the sake of comparison, both LC-KMP  and vanilla KMP are employed to generate the adapted CoM trajectories. As can be seen from Fig.~\ref{fig:loco:ada} (third - sixth plots), both LC-KMP (solid yellow curves) and vanilla KMP (dashed green curves) are capable of adapting trajectories towards various desired points (depicted by circles).  

In addition, evaluations of $a_x  x+ b_x \dot{x}$ and $a_y  y+ b_y \dot{y}$ are provided in Fig.~\ref{fig:capture:region}, where LC-KMP (solid yellow curves) fulfills the constraints of capture regions in both $X$ and $Y$ directions, while vanilla KMP (dashed green curves) exceeds the capture region in $Y$ direction. 
Therefore, in contrast to  vanilla KMP, LC-KMP can indeed take into account additional linear constraints while performing learning and adaptations of demonstrations.
Snapshots of walking motions on the simulation platform \cite{muratore2017xbotcore} are illustrated in Fig.~\ref{fig:loco:simu}.

\section{Conclusions\label{sec:conclusion}}
We have developed a novel imitation learning framework, which is capable of addressing the learning and adaptation issues while considering additional linear constraints. This framework has been verified through several examples, comprising adapting 2D letters with (partial) motion limits, adapting 3D letter in different planes, as well as planning stable walking trajectories for a humanoid robot. 
In this paper, we only focus on learning time-driven trajectories. In fact, due to the kernel treatment, the proposed framework can be extended to the learning of trajectories with high-dimensional inputs. It is worth mentioning that, in our previous work \cite{huang2017}, KMP has been proven effective in a human-robot collaboration setting, where the user's hand positions (6D input) were used to drive the robot movement straightforwardly. Thus, it would be interesting to exploit the new framework LC-KMP in this direction.

\section*{Acknowledgement}
We thank Jiatao Ding from Wuhan university and Istituto Italiano di Tecnologia for his help on locomotion evaluations.

%\newpage
%\addtolength{\textheight}{-10.5cm}   % This command serves to balance the column lengths
\bibliographystyle{IEEEtran}
\bibliography{bibiography}

\begin{thebibliography}{10}
\providecommand{\url}[1]{#1}
\csname url@samestyle\endcsname
\providecommand{\newblock}{\relax}
\providecommand{\bibinfo}[2]{#2}
\providecommand{\BIBentrySTDinterwordspacing}{\spaceskip=0pt\relax}
\providecommand{\BIBentryALTinterwordstretchfactor}{4}
\providecommand{\BIBentryALTinterwordspacing}{\spaceskip=\fontdimen2\font plus
\BIBentryALTinterwordstretchfactor\fontdimen3\font minus
  \fontdimen4\font\relax}
\providecommand{\BIBforeignlanguage}[2]{{%
\expandafter\ifx\csname l@#1\endcsname\relax
\typeout{** WARNING: IEEEtran.bst: No hyphenation pattern has been}%
\typeout{** loaded for the language `#1'. Using the pattern for}%
\typeout{** the default language instead.}%
\else
\language=\csname l@#1\endcsname
\fi
#2}}
\providecommand{\BIBdecl}{\relax}
\BIBdecl

\bibitem{atkeson1997robot}
C.~G. Atkeson and S.~Schaal, ``Robot learning from demonstration,'' in
  \emph{ICML}, vol.~97, 1997, pp. 12--20.

\bibitem{schaal1999imitation}
S.~Schaal, ``Is imitation learning the route to humanoid robots?'' \emph{Trends
  in cognitive sciences}, vol.~3, no.~6, pp. 233--242, 1999.

\bibitem{billard2008robot}
A.~Billard, S.~Calinon, R.~Dillmann, and S.~Schaal, ``Robot programming by
  demonstration,'' \emph{Springer handbook of robotics}, pp. 1371--1394, 2008.

\bibitem{ijspeert}
A.~J. Ijspeert, J.~Nakanishi, H.~Hoffmann, P.~Pastor, and S.~Schaal,
  ``Dynamical movement primitives: learning attractor models for motor
  behaviors,'' \emph{Neural computation}, vol.~25, no.~2, pp. 328--373, 2013.

\bibitem{calinon2014task}
S.~Calinon, D.~Bruno, and D.~G. Caldwell, ``A task-parameterized probabilistic
  model with minimal intervention control,'' in \emph{2014 IEEE International
  Conference on Robotics and Automation (ICRA)}.\hskip 1em plus 0.5em minus
  0.4em\relax IEEE, 2014, pp. 3339--3344.

\bibitem{huang2017}
Y.~Huang, L.~Rozo, J.~Silv{\'e}rio, and D.~G. Caldwell, ``Kernelized movement
  primitives,'' \emph{The International Journal of Robotics Research}, vol.~38,
  no.~7, pp. 833--852, 2019.

\bibitem{ude2010task}
A.~Ude, A.~Gams, T.~Asfour, and J.~Morimoto, ``Task-specific generalization of
  discrete and periodic dynamic movement primitives,'' \emph{IEEE Transactions
  on Robotics}, vol.~26, no.~5, pp. 800--815, 2010.

\bibitem{stulp2011learning}
F.~Stulp, E.~Theodorou, J.~Buchli, and S.~Schaal, ``Learning to grasp under
  uncertainty,'' in \emph{2011 IEEE International Conference on Robotics and
  Automation}.\hskip 1em plus 0.5em minus 0.4em\relax IEEE, 2011, pp.
  5703--5708.

\bibitem{huang2016}
Y.~Huang, D.~B{\"u}chler, O.~Ko{\c{c}}, B.~Sch{\"o}lkopf, and J.~Peters,
  ``Jointly learning trajectory generation and hitting point prediction in
  robot table tennis,'' in \emph{2016 IEEE-RAS 16th International Conference on
  Humanoid Robots (Humanoids)}.\hskip 1em plus 0.5em minus 0.4em\relax IEEE,
  2016, pp. 650--655.

\bibitem{howard2009novel}
M.~Howard, S.~Klanke, M.~Gienger, C.~Goerick, and S.~Vijayakumar, ``A novel
  method for learning policies from variable constraint data,''
  \emph{Autonomous Robots}, vol.~27, no.~2, pp. 105--121, 2009.

\bibitem{armesto2017learning}
L.~Armesto, V.~Ivan, J.~Moura, A.~Sala, and S.~Vijayakumar, ``Learning
  constrained generalizable policies by demonstration.'' in \emph{Robotics:
  Science and systems}, 2017.

\bibitem{ratliff}
N.~Ratliff, M.~Zucker, J.~A. Bagnell, and S.~Srinivasa, ``Chomp: Gradient
  optimization techniques for efficient motion planning,'' in \emph{2009 IEEE
  International Conference on Robotics and Automation}.\hskip 1em plus 0.5em
  minus 0.4em\relax IEEE, 2009, pp. 489--494.

\bibitem{chaumette2001redundancy}
F.~Chaumette and {\'E}.~Marchand, ``A redundancy-based iterative approach for
  avoiding joint limits: Application to visual servoing,'' \emph{IEEE
  Transactions on Robotics and Automation}, vol.~17, no.~5, pp. 719--730, 2001.

\bibitem{calinon2007}
S.~Calinon, F.~Guenter, and A.~Billard, ``On learning, representing, and
  generalizing a task in a humanoid robot,'' \emph{IEEE Transactions on
  Systems, Man, and Cybernetics, Part B (Cybernetics)}, vol.~37, no.~2, pp.
  286--298, 2007.

\bibitem{muhlig2009task}
M.~Muhlig, M.~Gienger, S.~Hellbach, J.~J. Steil, and C.~Goerick, ``Task-level
  imitation learning using variance-based movement optimization,'' in
  \emph{2009 IEEE International Conference on Robotics and Automation}.\hskip
  1em plus 0.5em minus 0.4em\relax IEEE, 2009, pp. 1177--1184.

\bibitem{huang2018towards}
Y.~Huang, J.~Silv{\'e}rio, and D.~G. Caldwell, ``Towards minimal intervention
  control with competing constraints,'' in \emph{2018 IEEE/RSJ International
  Conference on Intelligent Robots and Systems (IROS)}.\hskip 1em plus 0.5em
  minus 0.4em\relax IEEE, 2018, pp. 733--738.

\bibitem{cohn}
D.~A. Cohn, Z.~Ghahramani, and M.~I. Jordan, ``Active learning with statistical
  models,'' \emph{Journal of artificial intelligence research}, vol.~4, pp.
  129--145, 1996.

\bibitem{calinon2016}
S.~Calinon, ``A tutorial on task-parameterized movement learning and
  retrieval,'' \emph{Intelligent Service Robotics}, vol.~9, no.~1, pp. 1--29,
  2016.

\bibitem{paraschos}
A.~Paraschos, C.~Daniel, J.~R. Peters, and G.~Neumann, ``Probabilistic movement
  primitives,'' in \emph{Advances in neural information processing systems},
  2013, pp. 2616--2624.

\bibitem{saunders}
C.~Saunders, A.~Gammerman, and V.~Vovk, ``Ridge regression learning algorithm
  in dual variables,'' in \emph{Proceedings of the Fifteenth International
  Conference on Machine Learning}.\hskip 1em plus 0.5em minus 0.4em\relax
  Morgan Kaufmann Publishers Inc., 1998, pp. 515--521.

\bibitem{murphy}
K.~P. Murphy, \emph{Machine learning: a probabilistic perspective}.\hskip 1em
  plus 0.5em minus 0.4em\relax MIT press, 2012.

\bibitem{petersen2008matrix}
K.~B. Petersen, M.~S. Pedersen \emph{et~al.}, ``The matrix cookbook,''
  \emph{Technical University of Denmark}, vol.~7, no.~15, p. 510, 2008.

\bibitem{alessio2009survey}
A.~Alessio and A.~Bemporad, ``A survey on explicit model predictive control,''
  in \emph{Nonlinear model predictive control}.\hskip 1em plus 0.5em minus
  0.4em\relax Springer, 2009, pp. 345--369.

\bibitem{rawlings2000tutorial}
J.~B. Rawlings, ``Tutorial overview of model predictive control,'' \emph{IEEE
  control systems magazine}, vol.~20, no.~3, pp. 38--52, 2000.

\bibitem{pratt2006capture}
J.~Pratt, J.~Carff, S.~Drakunov, and A.~Goswami, ``Capture point: A step toward
  humanoid push recovery,'' in \emph{2006 6th IEEE-RAS international conference
  on humanoid robots}.\hskip 1em plus 0.5em minus 0.4em\relax IEEE, 2006, pp.
  200--207.

\bibitem{koolen2012capturability}
T.~Koolen, T.~De~Boer, J.~Rebula, A.~Goswami, and J.~Pratt,
  ``Capturability-based analysis and control of legged locomotion, part 1:
  Theory and application to three simple gait models,'' \emph{The International
  Journal of Robotics Research}, vol.~31, no.~9, pp. 1094--1113, 2012.

\bibitem{muratore2017xbotcore}
L.~Muratore, A.~Laurenzi, E.~M. Hoffman, A.~Rocchi, D.~G. Caldwell, and N.~G.
  Tsagarakis, ``Xbotcore: A real-time cross-robot software platform,'' in
  \emph{2017 First IEEE International Conference on Robotic Computing
  (IRC)}.\hskip 1em plus 0.5em minus 0.4em\relax IEEE, 2017, pp. 77--80.

\bibitem{ding2019nonlinear}
J.~Ding, X.~Xiao, and N.~G. Tsagarakis, ``Nonlinear optimization of step
  duration and step location,'' in \emph{IEEE/RSJ International Conference on
  Intelligent Robots and Systems}, 2019, to appear.

\end{thebibliography}

\end{document}